\documentclass[10pt,twocolumn,letterpaper]{article}

\usepackage[pagenumbers]{cvpr} 

\usepackage{booktabs}   
\usepackage{multirow}   
\usepackage{graphicx}   

\definecolor{cvprblue}{rgb}{0.21,0.49,0.74}
\usepackage[pagebackref,breaklinks,colorlinks,allcolors=cvprblue]{hyperref}
\usepackage{tikz}
\usepackage{listings}
\usepackage{xcolor}

\usetikzlibrary{shapes.geometric, arrows.meta, positioning, calc}

\title{TAG-MoE: \underline{T}ask-\underline{A}ware \underline{G}ating for Unified Generative \underline{M}ixture-\underline{o}f-\underline{E}xperts}

\author{
Yu Xu\textsuperscript{1,2†} \quad
Hongbin Yan\textsuperscript{1} \quad
Juan Cao\textsuperscript{1} \quad
Yiji Cheng\textsuperscript{2} \quad
Tiankai Hang\textsuperscript{2} \quad \\
Runze He\textsuperscript{2} \quad
Zijin Yin\textsuperscript{2} \quad
Shiyi Zhang\textsuperscript{2} \quad
Yuxin Zhang\textsuperscript{1} \quad
Jintao Li\textsuperscript{1} \quad \\
Chunyu Wang\textsuperscript{2‡} \quad
Qinglin Lu\textsuperscript{2} \quad
Tong-Yee Lee\textsuperscript{3} \quad
Fan Tang\textsuperscript{1§} \\
\textsuperscript{1}University of Chinese Academy of Sciences \quad
\textsuperscript{2}Tencent Hunyuan \quad
\textsuperscript{3}National Cheng-Kung University \\
\url{https://yuci-gpt.github.io/TAG-MoE/}
}

\begin{document}

\twocolumn[{
\renewcommand\twocolumn[1][]{#1}
\maketitle
\begin{center}
    \centering
    \vspace*{-.8cm}
    \includegraphics[width=.99\textwidth]{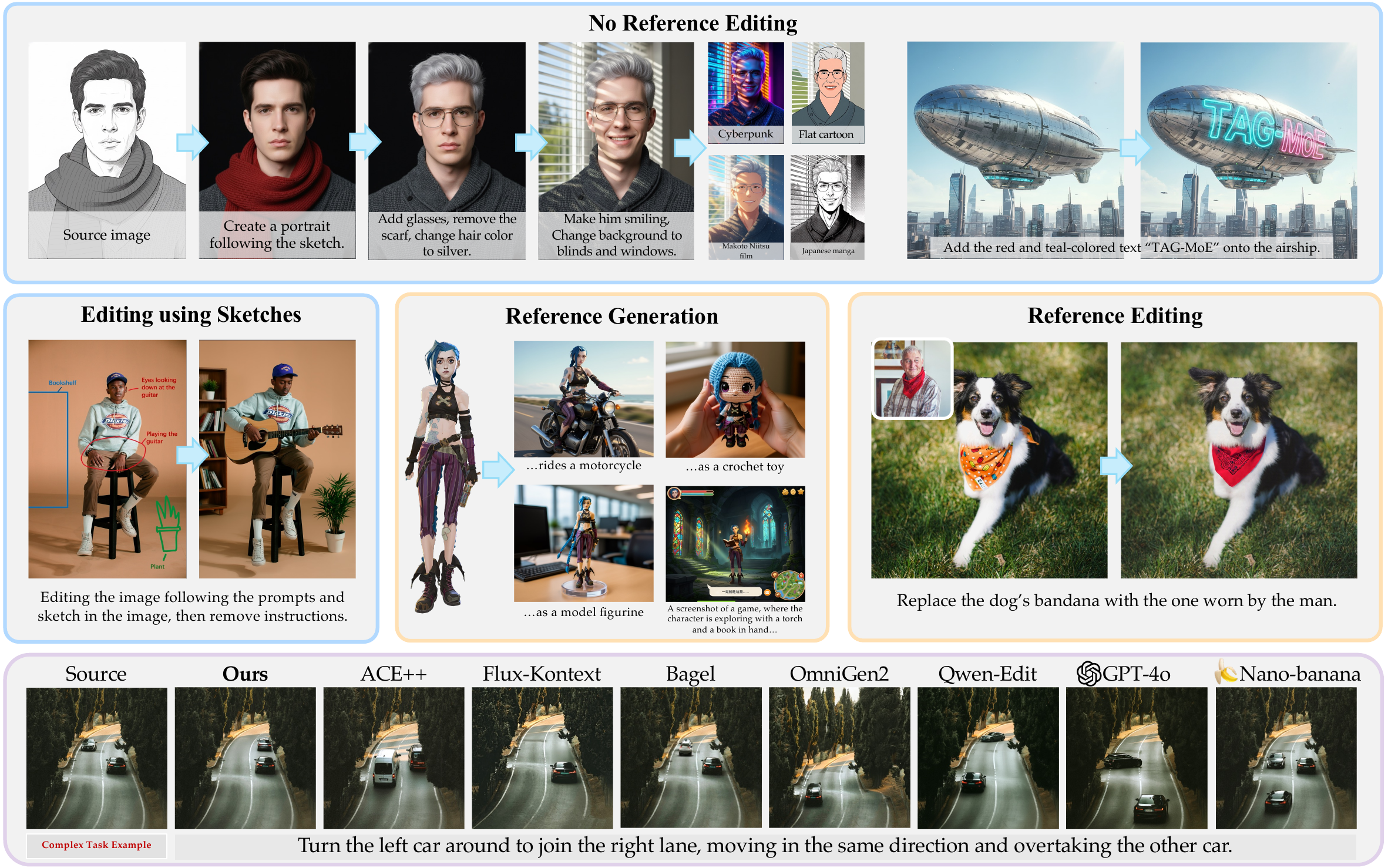}
    \captionof{figure}{We present \textbf{TAG-MoE}, by injecting high-level task semantic intent into the local routing decisions of the MoE gating network, we enabling the diffusion transformer model to handle diverse generative tasks.}
\label{fig:teaser}
\end{center}
}]

\renewcommand{\thefootnote}{}
\footnotetext{
\hspace{-2.05em}
\raggedright
\textsuperscript{†} Work done during internship at Tencent Hunyuan. \\
\textsuperscript{‡} Project leader. \\
\textsuperscript{§} Corresponding author. tfan.108@gmail.com
}

\maketitle
\begin{abstract}

Unified image generation and editing models suffer from severe task interference in dense diffusion transformers architectures, where a shared parameter space must compromise between conflicting objectives (e.g., local editing v.s. subject-driven generation). While the sparse Mixture-of-Experts (MoE) paradigm is a promising solution, its gating networks remain task-agnostic, operating based on local features, unaware of global task intent. This task-agnostic nature prevents meaningful specialization and fails to resolve the underlying task interference.
In this paper, we propose a novel framework to inject semantic intent into MoE routing. We introduce a Hierarchical Task Semantic Annotation scheme to create structured task descriptors (e.g., scope, type, preservation). We then design Predictive Alignment Regularization to align internal routing decisions with the task's high-level semantics. This regularization evolves the gating network from a task-agnostic executor to a dispatch center. Our model effectively mitigates task interference, outperforming dense baselines in fidelity and quality, and our analysis shows that experts naturally develop clear and semantically correlated specializations.

\end{abstract}

\section{Introduction}
\label{sec:intro}
The field of visual synthesis is rapidly converging toward unified image generation and editing models~\cite{hurst2024gpt, comanici2025gemini, labs2025flux, xu2026beyond}, frameworks designed to consolidate disparate image manipulation tasks—from subject customization and style transfer to high-fidelity inpainting and instruction-based editing—into a single, robust system with the help of large-scale, dense Diffusion Transformers (DiT). 

While promising efficiency, this unification is critically bottlenecked by severe task interference.
The shared parameter space must simultaneously execute inherently contradictory objectives: local editing demands precise content preservation, while subject-driven generation requires expressive diversity and novel synthesis.
This fundamental conflict forces the network toward a ``mediocre compromise solution,'' preventing the necessary representational specialization and ultimately degrading performance across the spectrum of user intents.

To overcome the scalability~\cite{fei2024scaling} and capacity~\cite{zheng2025dense2moe} limits of dense DiT, the sparse Mixture-of-Experts (MoE) paradigm is adopted to dramatically expand model capacity with manageable inference costs of large-scale generative models. 
However, these efforts mainly focus on single, general-purpose image generation tasks, and have not (and do not need to) account for the complex task diversity within the unified generation framework. 
Applying standard MoE to the heterogeneous unified domain introduces a critical architectural failure: the \textbf{task-agnostic} nature of conventional gating networks. 
Standard routers rely solely on local token features, remaining entirely oblivious to the high-level, global task intent (e.g., ``identity preservation'' or ``style modification''). 
This profound information gap between the local gate and the global objective leads to spontaneous, inefficient expert specialization, fundamentally failing to structurally disentangle multi-task interference.  
\textit{How to inject the high-level, global task semantics into the local MoE routing mechanism to enable task-aware specialization remains an open challenge.}

In this study, we propose \textbf{TAG-MoE}, a task-aware gating network for unified image generation and editing. 
First, to provide a structured unified task representation, we introduce a \textbf{hierarchical task semantic annotation} scheme, by decomposing specific generative task into a multi-faceted descriptor, capturing the operational \textit{scope} (e.g., {local/global editing}), the semantic \textit{type} (e.g., {attribute/action editing}), and essential \textit{preservation constraints} (e.g., {identity/style preservation}). 
Such structured representations provides the necessary rich supervisory signal previously missing.
Furthermore, we propose a novel training framework founded on the principle that semantically similar generation tasks evokes similar expert usage patterns. 
To enforce this, we design an innovative \textbf{predictive alignment regularization} to correlate the high-level task semantic intent with the underlying routing decisions. 
Such regularization serves as a bridge to compel the model's internal routing strategy to become predictive of the task's macro-semantics, injecting global semantic intent into the local routing mechanism, leading the gating network to evolve from a task-agnostic executor into an aware, intelligent dispatch center. 
Experiments on unified image generation benchmarks ICE-Bench, image editing benchmark EmuEdit and GEdit, subject-driven generation benchmark DreamBench++ and OmniContext indicate that our method achieves the best overall performance.
Our primary contributions are summarized as follows:
\begin{enumerate}
    \item We propose a novel task-aware sparse MoE framework and successfully apply it to Diffusion Transformer-based unified image generation and editing tasks.
    \item We introduce a hierarchical task semantic annotation scheme and a corresponding predictive alignment regularization that, together, effectively resolve the task-agnostic of the MoE gate by aligning its routing strategy with the task's semantic intent.
    \item By successfully mitigating task interference, our model achieves SOTA overall performance against open-source baselines across five comprehensive benchmarks.
\end{enumerate}

\section{Related Work}
\label{sec:related}
\subsection{Unified Image Generation and Editing}
Recent efforts in unified image generation aim to build single models capable of handling a broad range of image manipulation tasks, moving beyond specialized, task-specific approaches~\cite{xu2025attribute, xu2025context, dai2025latent, dai2026omni2sound}. Early methods treat the problem as a sequence-to-sequence task, concatenating text, source, and target image tokens for large transformers~\cite{xiao2025omnigen, han2025ace, fu2025univg}.
Subsequent works refine input representations and architectures to improve multimodal conditioning. Methods such as UniReal~\cite{chen2025unireal} and RealGeneral~\cite{lin2025realgeneral} introduces trainable index, subject, and condition embeddings to enhance alignment, while Flux-Kontext~\cite{labs2025flux} employes 3D rotary positional encodings to distinguish source from target images. Architectural innovations include dual-branch models that decouple subject and background processing~\cite{li2025blobctrl}, channel-wise concatenation to preserve contextual signals~\cite{mao2025ace++}, and the integration of auxiliary MLLMs or transformers for improved scene understanding~\cite{deng2025emerging, wu2025omnigen2, song2025query, yang2025magic}, albeit with increased complexity and compute.

Despite these advances, current unified models overlook a central challenge: the inherent conflict between the objectives of different image-to-image tasks. Editing tasks~\cite{wang2025ladb, xu2024headrouter, zhu2025kv, he2026re, yin2026generative} (e.g., style transfer, object removal) require precise regional preservation while modifying others, whereas customization tasks~\cite{ruiz2023dreambooth, xu2025b4m, wang2026geodesicnvs, he2024customize} (e.g., subject-driven generation) demand strong identity consistency across new contexts. Without explicitly modeling these distinct—and often competing—requirements, existing approaches struggle to adaptively serve the full spectrum of user intents, limiting their practical robustness and generalization.

\subsection{Image Generation with Mixture of Experts}
The MoE paradigm increases model capacity by routing inputs to specialized sub-networks, or “experts,” avoiding a proportional rise in per-sample computation. Its success in large language models has motivated adoption in visual generation: pioneering works such as DiT-MoE~\cite{fei2024scaling}, and scaled variants like HunyuanImage-3.0~\cite{cao2025hunyuanimage} and Dense2MoE~\cite{zheng2025dense2moe}, show that sparse expert architectures can enhance the expressiveness of diffusion transformers.
Extending MoE to image editing, ICEdit~\cite{zhang2025context} integrates LoRA-based MoE modules into attention blocks. However, purely data-driven routing is fundamentally limited: task-agnostic routers cannot resolve conflicts between heterogeneous tasks (e.g., editing vs. customization), and the restricted capacity of LoRA experts hampers learning multi-task behaviors.
Our approach overcomes these limitations by introducing task-aware expert routing. We condition the gating mechanism on learnable embeddings corresponding to specific task categories, enabling dynamic selection of the most relevant experts. This mitigates inter-task conflicts, promotes effective specialization, and achieves superior performance across diverse image-to-image tasks while maintaining the efficiency of the MoE framework.
\section{Method}
\label{sec:method}
\begin{figure*}[!h]
  \centering
  \includegraphics[width=1\linewidth]{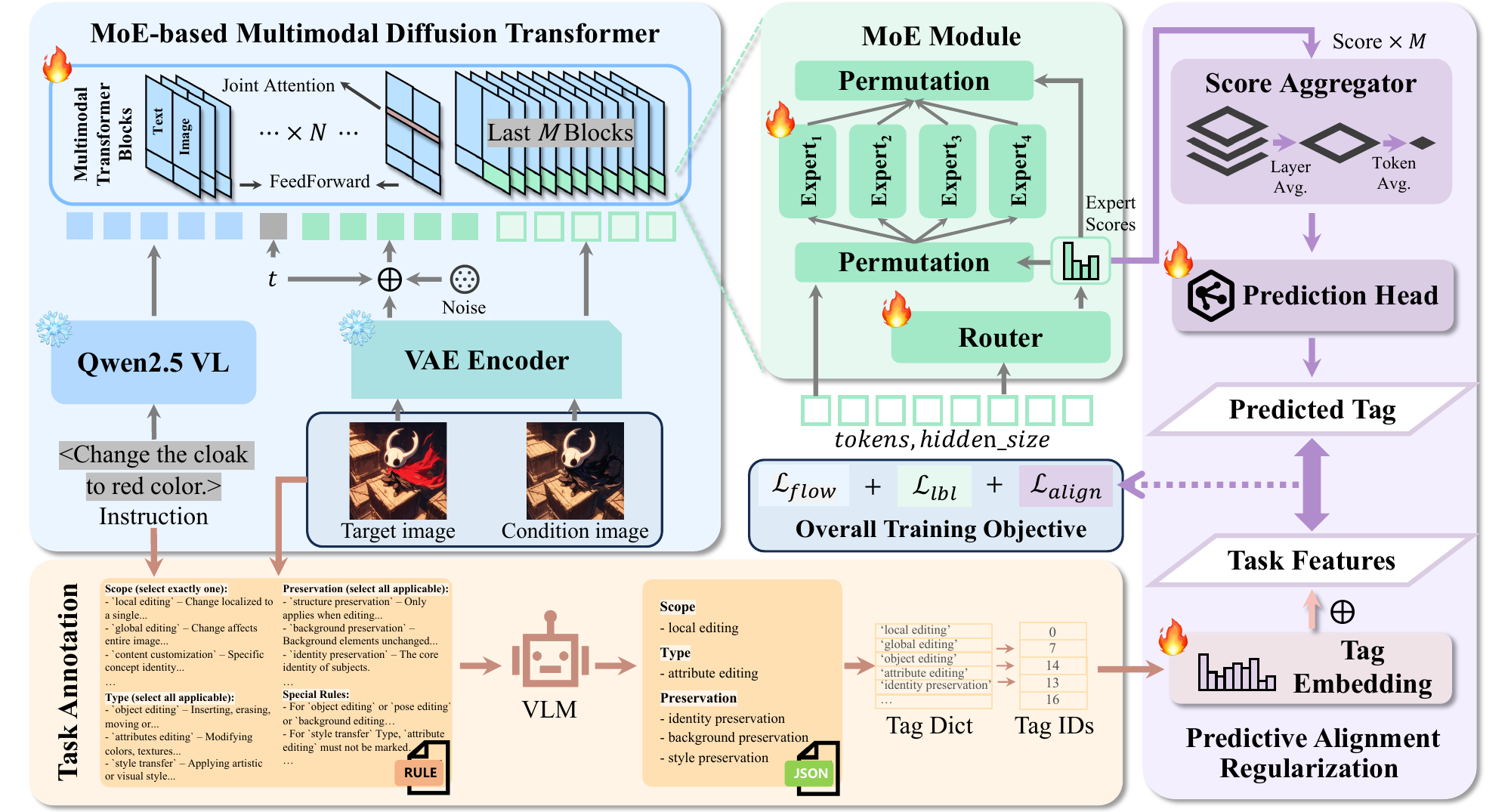}
  \caption{Pipeline of our method. TAG-MoE consists of: (1) A \textbf{MM-DiT with MoE} layers; (2) A \textbf{Hierarchical Task Semantic Annotation} that labels training data with atomic task descriptors; (3) A novel Semantic-Aligned Router explicitly aligns MoE routing behavior with task semantics through \textbf{Predictive Alignment Regularization}.}

  \label{fig:pipeline}
\end{figure*}
Our unified framework (Fig.~\ref{fig:pipeline}) employs a Multimodal Diffusion Transformer (MM-DiT) with MoE layers for efficient, dynamic task handling (\S\ref{ssec:moe_arch}). We introduce hierarchical task semantic annotation (\S\ref{ssec:tag_pipeline}) and a novel semantic-aligned router (\S\ref{ssec:sem_router}). This router guides the MoE's specialization by aligning its routing decisions with these explicit task semantics in an interpretable manner .

\subsection{MoE-based Multimodal Diffusion Transformer}
\label{ssec:moe_arch}
Building upon an MM-DiT architecture, our approach processes diverse inputs within a unified token sequence framework. To interpret user instructions, we employ a powerful pre-trained Multimodal Large Language Model (MLLM) to encode the input text $c_{text}$ into a sequence of text embeddings $C$. Separately, a pre-trained VAE encoder $\mathcal{E}$ maps both the conditional image $I_c$ and the target image $I_0$ into latent representations, $z_c$ and $z_0$. During training, Gaussian noise is sampled and added to the $z_0$ to produce a noisy version $z_t$. Both $z_c$ and $z_t$ are then patchified into sequences of visual tokens. Finally, the complete input to our MM-DiT is a single sequence formed by concatenating the text embeddings $C$, the image tokens from $z_c$, the image tokens from the noisy target latent $z_t$, and a timestep embedding~\cite{peebles2023scalable}.

We replace the feed-forward networks (FFNs) of the image stream in diffusion transformer blocks with MoE layers. This leverages sparse activation to significantly increase model capacity at a fixed activation parameter, enabling superior performance over dense models with a comparable budget. We only implement MoE layers in the later transformer blocks as high-level semantic synthesis in these deeper layers benefits most from the increased capacity~\cite{rajbhandari2022deepspeed, deepseekai2024deepseekv3technicalreport}.
The MoE layer consists of a set of $N$ expert networks $E$ and a gating network $\mathcal{G}$. The gating network $\mathcal{G}$ maps each input token to a probability distribution over the $N$ experts, thereby determining their top $k$ selections $\mathcal{T}\subseteq\left\{E_1,\dots,E_N\right\}$. The output is a weighted sum of the activated experts' outputs:

\begin{equation}
    \text{MoE}(x) = \sum_{E_i\in\mathcal{T}(x)} \mathcal{G}(x)_i \cdot E_i(x).
\end{equation}
This MoE-enhanced architecture is trained end-to-end using a Flow Matching objective.

\subsection{Hierarchical Task Semantic Annotation}
\label{ssec:tag_pipeline}
To train a unified model that supports a broad range of generation and editing tasks, a structured representation of task semantics is essential. A single coarse label (e.g., ``edit'') cannot capture user intent. For example, ``change the background to a beach'' and ``make the person smile' are both edits but require fundamentally different behaviors and preservation constraints.
To address this, we introduce a three-tier annotation scheme that provides each training instance (source image, instruction, target image) with a rich semantic descriptor: Scope - the task’s operational nature and spatial extent (e.g., global editing, local editing, content customization). Type — the semantic category of the manipulation (e.g., object editing, style transfer, attribute editing). Preservation — the invariants that must remain unchanged (e.g., identity, background, structure preservation).

An automated pipeline utilizing Qwen-VL~\cite{bai2025qwen2} is established to analyze training triplets. It involves providing definitions of a three-tier system and instructing Qwen-VL to output atomic tags. The rule set is continuously refined to maintain consistency and semantic quality.

For instance, the task ``Make the person in the photo wear sunglasses'' would be annotated with tags such as ``Scope: local editing; Type: object editing; Preservation: identity preservation, background preservation, style preservation''. This rich set of atomic tags forms the basis for our semantic representation. 

\par\textbf{Inference Stage.} This hierarchical annotation scheme is exclusively used for training. During the inference stage, these ground-truth tags are no longer required. Instead, as a lightweight pre-processing step, we pass the user's raw instruction $c_{text}$ and the source image $I_c$ to a VLM (e.g., Qwen-VL~\cite{bai2025qwen2}). The VLM performs instruction rewriting, analyzing the image and text to generate a more detailed, descriptive prompt. This enriched prompt is then encoded as the text embedding $C$ and fed into the MM-DiT.

\subsection{Semantic-Aligned Gating Network}
\label{ssec:sem_router}
We design a novel semantic-aligned gating network to force the model's internal routing strategy (encoded as a routing signature ``$\mathbf{g}$'') to predict the task's macroscopic semantics (encoded as a semantic embedding ``$\mathbf{s}$''). This predictive alignment serves as a bridge, connecting local routing decisions with global task intent. Our mechanism comprises three key components: (1) construction of the global semantic embedding $\mathbf{s}$; (2) construction of the aggregated routing signature $\mathbf{g}$; and (3) the predictive alignment loss $\mathcal{L}_{align}$.

\subsubsection{Global Semantic Embedding}
Based on the hierarchical task semantic annotation described in \S\ref{ssec:tag_pipeline}, we first define a global vocabulary $\mathcal{V}$ containing all $K$ atomic tags (e.g., ``local editing'', ``identity preservation''). We instantiate a learnable tag embedding matrix $\mathbf{W}_{tag} \in \mathbb{R}^{K \times D}$ for this vocabulary, where $D$ is the model's hidden dimension.
For a given training sample, its associated tags form a set $T_p \subseteq \mathcal{V}$ (e.g., $T_p = \{$``local editing'', ``face preservation''$\}$). To convert this variable-sized set $T_p$ into a fixed-dimension vector $\mathbf{s}$, we first retrieve the corresponding embedding vector $\mathbf{e}_t = \mathbf{W}_{tag}[\text{index}(t)]$ for each tag $t \in T_p$, and then aggregate them via element-wise summation.
This constructs the global semantic embedding $\mathbf{s}$, which represents the ``macro-level semantic ground truth'':
\begin{equation}
\mathbf{s} = \sum_{t \in T_p} \mathbf{W}_{tag}[\text{index}(t)].
\label{eq:semantic_embedding}
\end{equation}
This vector $\mathbf{s} \in \mathbb{R}^{D}$ is permutation-invariant, meaning the order of tags does not affect the final representation. It serves as the structured supervisory signal for our subsequent alignment loss.

\subsubsection{Aggregated Routing Signature}
Correspondingly, we require a vector to represent the internal routing strategy the model actually employs for the current sample. The gating network $\mathcal{G}$ (see \S\ref{ssec:moe_arch}) generates routing scores $S_{l,t} \in \mathbb{R}^{N}$ for each token $t$ in each of the $L$ MoE layers, where $N$ is the number of experts.

To obtain a single vector representing the expert usage pattern for the entire sample, we design an aggregated routing signature $\mathbf{g}$. First, we average the routing scores across all $L$ MoE layers to get a per-token average score $\bar{S}_t = \frac{1}{L} \sum_{l=1}^{L} S_{l,t}$. Next, we apply mean pooling over the sequence (token) dimension to get the final signature $\mathbf{g} \in \mathbb{R}^{N}$:
\begin{equation}
\mathbf{g} = \frac{1}{T} \sum_{t=1}^{T} \bar{S}_t 
= \frac{1}{T \cdot L} \sum_{t=1}^{T} \sum_{l=1}^{L} S_{l,t}.
\label{eq:routing_signature}
\end{equation}
This vector $\mathbf{g}$ encodes which experts are activated on average to process the sample, capturing its \textit{de facto} internal routing policy.

\subsubsection{Predictive Alignment Regularization}
We now have two vectors: $\mathbf{s} \in \mathbb{R}^{D}$, representing what the task \textit{should be}, and $\mathbf{g} \in \mathbb{R}^{N}$, representing what the model \textit{actually do}. To align them, we introduce a lightweight prediction head $\mathcal{H}_{pred}$ (a two-layer MLP), to project the aggregated routing signature $\mathbf{g}$ from the expert space $\mathbb{R}^{N}$ into the semantic space $\mathbb{R}^{D}$, yielding a predicted semantic embedding $\hat{\mathbf{s}} = \mathcal{H}_{pred}(\mathbf{g})$.

We force the routing strategy to predict the task semantics by minimizing the cosine similarity loss between $\hat{\mathbf{s}}$ and $\mathbf{s}$. This is our Predictive Alignment Loss $\mathcal{L}_{align}$:
\begin{equation}
\mathcal{L}_{align} = 1 - \text{sim}(\hat{\mathbf{s}}, \mathbf{s}) = 1 - \frac{\hat{\mathbf{s}} \cdot \mathbf{s}}{|\hat{\mathbf{s}}| |\mathbf{s}|}.
\label{eq:align_loss}
\end{equation}
Minimizing $\mathcal{L}_{align}$ trains the parameters of $\mathcal{H}_{pred}$ and, more importantly, backpropagates the gradient through $\mathbf{g}$ to the gating networks $\mathcal{G}$ of all MoE layers. This compels $\mathcal{G}$ to evolve from a task-agnostic executor into a semantic-aware scheduler: it must learn to route tokens intelligently, such that the resulting aggregate signature $\mathbf{g}$ contains sufficient information to predict the global task $\mathbf{s}$.

\subsubsection{Overall Training Objective}
Our proposed $\mathcal{L}_{align}$ is an auxiliary loss that complements the model's primary objective. The final overall loss $\mathcal{L}_{total}$ is a weighted sum of the main generation loss (e.g., $\mathcal{L}_{flow}$), the standard MoE load balancing loss $\mathcal{L}_{lbl}$, and our semantic alignment loss $\mathcal{L}_{align}$:
\begin{equation}
\mathcal{L}_{total} = \mathcal{L}_{flow} + \lambda_{lbl} \mathcal{L}_{lbl} + \lambda_{align} \mathcal{L}_{align},
\label{eq:total_loss}
\end{equation}
where $\lambda_{lbl}$ and $\lambda_{align}$ are hyperparameters that balance the contribution of each loss term.

\subsection{Dataset Construction}
Our model is trained on a large-scale, diverse dataset comprising both publicly available and proprietary in-house data, totaling over 11 million samples. This hybrid approach ensures broad coverage across the unified task space. The public portion (2.2M samples) is compiled from established benchmarks, including InstructP2P~\cite{brooks2023instructpix2pix}, UltraEdit~\cite{zhao2024ultraedit}, and OmniEdit~\cite{wei2024omniedit} for universal instructive editing, supplemented by VTON-HD~\cite{choi2021viton} for virtual try-on tasks and Ominicontrol~\cite{tan2025ominicontrol} for subject driven generation.

Our proprietary in-house dataset is meticulously constructed using a multi-stage pipeline to cover a wide spectrum of specialized tasks. First, we source pristine images from large-scale public datasets. Next, we employ large language models (e.g., GPT-4o~\cite{openai2025gpt4o}) to generate a vast array of diverse editing and generation instructions for these images. To obtain high-quality target images, we utilize a combination of specialist and generalist models: for instance, specialist models like ControlNet~\cite{zhang2023adding} are used for ``Control generation'' tasks, while powerful generalist models (e.g., Flux-Kontext~\cite{labs2025flux}, Qwen-Edit~\cite{wu2025qwen}, and SeedEdit~\cite{wang2025seededit}) are employed for a broad range of edits. Following the methodology of UniReal~\cite{chen2025unireal}, we also process video frames to create dynamic editing datasets (e.g., for pose/view changes). Finally, to enhance robustness and quality, we systematically augment the data by constructing corresponding inverse tasks and instructions (e.g., pairing ``object addition'' with ``object removal''), which significantly improves generative fidelity.
\section{Experiments}
\label{sec:experiment}
\subsection{Implenentation Details}
Our model is based on Qwen-Image T2I model~\cite{wu2025qwen}, we integrate the MoE layers by replacing the standard FFNs of the image stream in the final 10 layers of our diffusion transformer. Each MoE layer consists of four experts, where each expert possesses an architecture identical to the original FFN it replaces. The gating network is implemented as a two-layer MLP, and we employ a top-1 routing strategy.

\subsection{Experiments Settings}

\paragraph{Baselines.}
We compare our method against three categories of SOTA baselines.
(1) Unified generation and editing methods for diverse image-to-image tasks, including ACE++~\cite{mao2025ace++}, Flux.1 Kontext~\cite{labs2025flux}, BAGEL~\cite{deng2025emerging}, OmniGen2~\cite{wu2025omnigen2}, Qwen-Edit~\cite{wu2025omnigen2} and DreamOmni2~\cite{xia2025dreamomni2}. 
We also include comparisons against product-level, closed-source models (e.g. GPT-4o~\cite{openai2025gpt4o} and Gemini-2.5-flash (aka. Nano-banana)~\cite{google2025nanobanana}, to contextualize our performance. However, our primary quantitative evaluation and main claims are benchmarked against open-source baselines.
(2) Specialized zero-shot instruction-based editing methods, including InstructPix2Pix~\cite{brooks2023instructpix2pix}, EmuEdit~\cite{sheynin2024emu}, MagicBrush~\cite{zhang2023magicbrush}, UltraEdit~\cite{zhao2024ultraedit}, ICEdit~\cite{pan2025ice}, and Step1X-Edit~\cite{liu2025step1x}.
(3) Specialized zero-shot subject-driven generation methods, including DreamO~\cite{mou2025dreamo}, OminiControl~\cite{tan2025ominicontrol} and UNO~\cite{wu2025less}.
\paragraph{Evaluation benchmarks.}
To comprehensively assess our model in the unified image generation and editing setting, we adopt ICE-Bench~\cite{pan2025ice} as our primary benchmark, as it is specifically designed for unified models and spans both diverse editing tasks and subject-driven generation. For more fine-grained evaluation, we further include specialized benchmarks: EmuEdit-Bench~\cite{sheynin2024emu} and GEdit-Bench~\cite{liu2025step1x} for detailed editing analysis, and DreamBench++~\cite{peng2024dreambench} together with OmniContext~\cite{wu2025omnigen2} to evaluate subject-driven generation performance.
\paragraph{Metrics.}

We employ a comprehensive set of metrics to evaluate both visual quality and task correctness. Aesthetic quality is assessed using a SigLip-based predictor. Consistency with the source image is measured via CLIP-src (for editing) and CLIP-ref (for subject-driven generation), while text alignment is captured by CLIP-cap.
For editing evaluation, we further use Qwen2-VL-72B~\cite{wang2024qwen2} to determine whether the instruction is correctly executed based on the source image, instruction, and output image, yielding the vllmqa score. For subject-driven tasks, we assess three key preservation dimensions: facial identity (Face-ref, using the buffalo model from InsightFace App~\cite{deepinsight2021insightface}), subject similarity (DINO-ref, via DINO~\cite{caron2021emerging}), and style fidelity (Style-ref, via CSD~\cite{somepalli2024measuring}).
All metrics not originally within the [-1, 1] range are normalized. For every metric reported, higher values indicate better performance. In the tables, the best results are highlighted \textbf{in bold}, and the second-best results are \underline{underlined}.

\begin{figure*}[!h]
  \centering
  \includegraphics[width=0.95\linewidth]{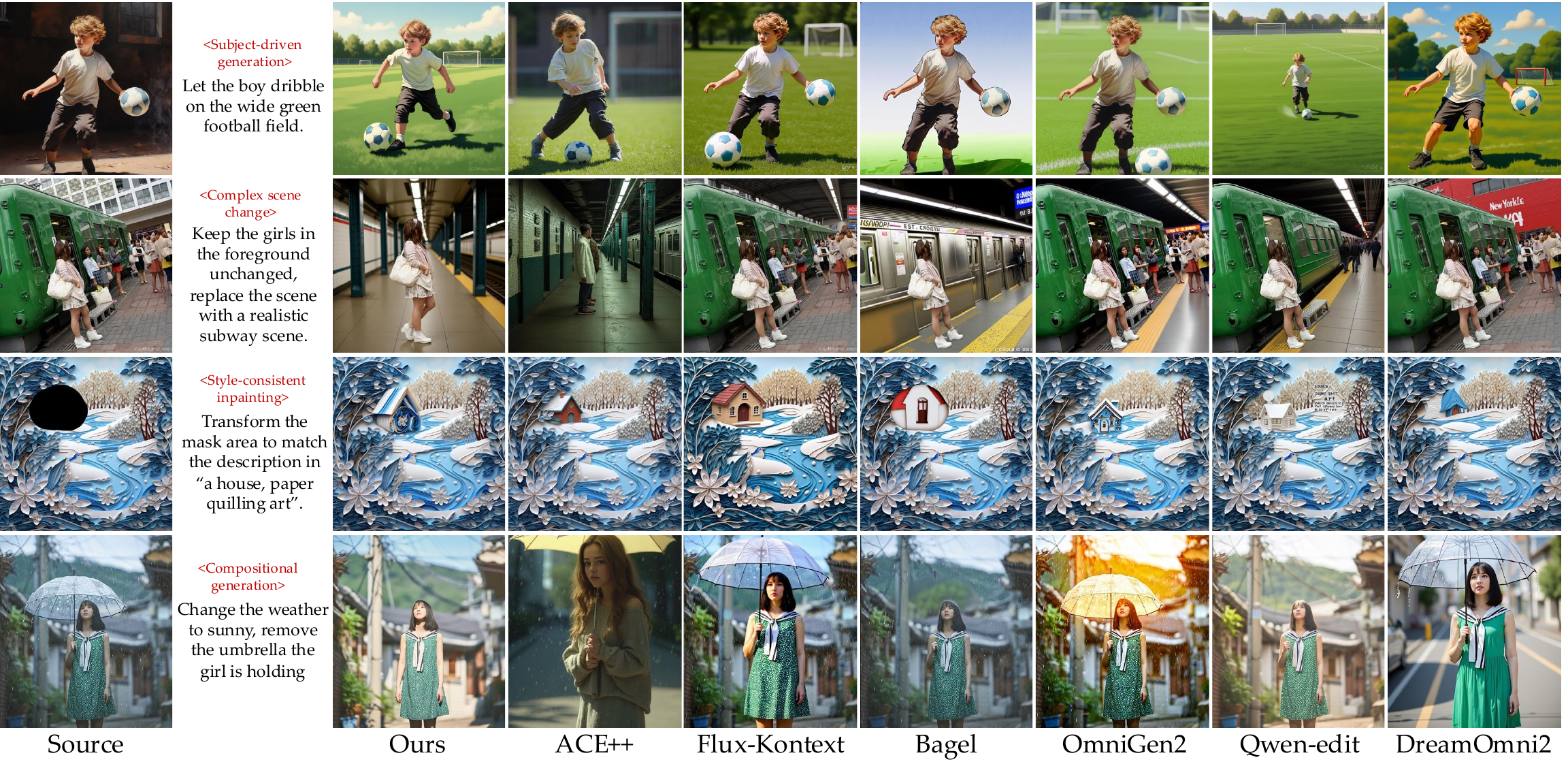}
  \caption{Qualitative comparison on diverse tasks.
  Our model successfully resolves complex task conflicts where baselines fail. 
  }
  \label{fig:main_compare}
\end{figure*}
\subsection{Quantitative Comparison}
\paragraph{Unified generation evaluation.}

We report the main results on ICE-Bench in Tab.~\ref{tab:main_compare}. Our method achieves the highest scores among all open-source baselines across three key metrics: aesthetic quality, CLIP-cap, and vllmqa. Notably, our CLIP-cap score not only surpasses all open-source competitors but also exceeds closed-source, product-level models such as GPT-4o and Gemini-2.5-flash, indicating stronger alignment with user instructions across diverse generation and editing tasks.
Although some baselines exhibit high source fidelity (e.g., DreamOmni2 on CLIP-src), our model attains a more favorable overall balance by excelling in instruction adherence and semantic alignment.

\begin{figure}[!h]
  \centering
  \includegraphics[width=0.95\linewidth]{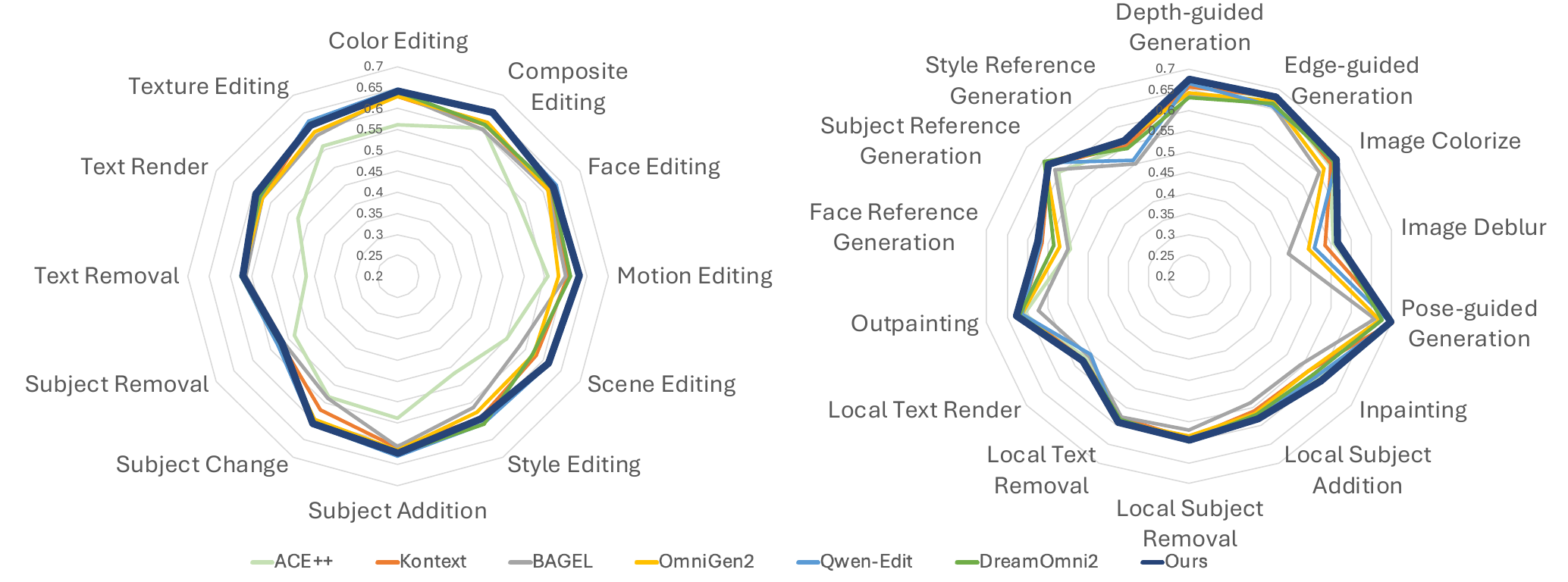}
  \caption{Comprehensive scores on different image editing and generation tasks. }
  \label{fig:chart}
\end{figure}

We further present a per-category breakdown over 26 task types on ICE-Bench, visualized in the radar charts in Fig.~\ref{fig:chart}. Our model achieves state-of-the-art performance in the vast majority of categories, demonstrating robust and well-balanced capability. DreamOmni2’s high reference-generation scores largely stem from copy-paste behavior on source subjects, which artificially inflates similarity metrics.

\begin{table}[t]
\centering
\small
\setlength{\tabcolsep}{3pt}
\renewcommand{\arraystretch}{1.1}
\resizebox{\columnwidth}{!}{
\begin{tabular}{l|ccccc}
\hline
\textbf{Method} &Aes.  & CLIP-src  & CLIP-cap & CLIP-ref & vllmqa \\ \hline
ACE++     &5.219 &0.851   &0.263 &0.713 &0.637  \\
Kontext      &5.165 &\underline{0.863}   &0.274 &0.728 &0.629  \\
BAGEL        &4.757 &0.863   &0.276 &0.687 &0.699 \\
OmniGen2  &5.238 &0.855   &\underline{0.279} &0.728 &\underline{0.787}  \\
Qwen-Edit  &\underline{5.358} &0.840   &\underline{0.279} &0.671 &0.774  \\
DreamOmni2   &5.188 &\textbf{0.866}   &0.268 &\textbf{0.739} &0.664  \\
Ours     & \textbf{5.399} & 0.857   & \textbf{0.282} & \underline{0.732} & \textbf{0.852}  \\
\hline
GPT-4o     &5.801 &0.823   &0.278 &0.693 &0.889   \\
Gemini-2.5-flash     &5.571 &0.879   &0.281 &0.724 &0.847   \\
\hline
\end{tabular}
}
\caption{Comparison results for unified tasks on ICE-Bench~\cite{pan2025ice} test sets. Open-source models are in the first block and close-source produce-level models are in the second block.}
\label{tab:main_compare}
\end{table}


\paragraph{Image editing evaluation}
We further evaluate our model against specialized zero-shot editing baselines on EmuEdit-bench~\cite{sheynin2024emu} and GEdit-bench~\cite{liu2025step1x}, with results shown in Tab.~\ref{tab:edit_compare}. (Note: Since EmuEdit is not open-source and only provides pre-generated outputs on its own benchmark, its performance on GEdit-bench is unavailable.)
Although our model does not achieve top-1 performance on every metric, it clearly leads on the most important indicator vllmqa achieving the highest scores on both benchmarks. This is particularly noteworthy because, unlike static CLIP similarity, vllmqa uses a powerful VLLM to evaluate the correctness of the executed instruction, offering a more intelligent and reliable measure of editing success. Our strong results on this metric underscore the model’s advanced instruction-following capability.

\begin{table}[ht]
\centering
\small
\setlength{\tabcolsep}{4pt}
\renewcommand{\arraystretch}{1.1}
\resizebox{\columnwidth}{!}{
\begin{tabular}{l|ccc|ccc}
\hline
\multirow{2}{*}{\textbf{Method}} &
\multicolumn{3}{c|}{\textbf{EmuEdit-bench}} &
\multicolumn{3}{c}{\textbf{GEdit-bench}} \\
\cline{2-7}
 & CLIP-src & CLIP-cap & vllmqa
 & CLIP-src & CLIP-cap & vllmqa \\
\hline
InsP2P  & 0.8589 & 0.2919 & 0.2507 & 0.8604 & 0.3192 & 0.3191 \\
EmuEdit  & 0.8854 & 0.3098 & 0.6253 & - & - & - \\
MagicBrush  & 0.8552 & 0.2951 & 0.4573 & 0.8068 & 0.3146 & 0.3783 \\
UltraEdit  & 0.8625 & 0.3075 & 0.3609 & 0.8459 & 0.3323 & 0.4605 \\
ICEdit  & 0.8912 & 0.3026 & 0.3609 & 0.9007 & 0.3283 & 0.4145 \\
Step1X-Edit  & 0.8845 & 0.3119 & 0.7893 & 0.8967 & 0.346 & \underline{0.8158} \\

\hline
ACE++  & 0.8367 & 0.2385 & 0.0606 & 0.8160 & 0.2518 & 0.0559 \\
Kontext  & \textbf{0.9091} & 0.3093 & 0.741 & 0.9190 & 0.3419 & 0.7303 \\
BAGEL  & 0.8565 & 0.3129 & 0.7989 & 0.8727 & 0.3470 & 0.7961 \\
OmniGen2  & 0.8932 & 0.3087 & 0.5978 & 0.8940 & 0.3373 & 0.6546 \\
Qwen-Edit  & 0.8832 & \textbf{0.3159} & \underline{0.9174} & 0.9104 & \textbf{0.3522} & 0.875 \\
DreamOmni2  & 0.9035 & 0.3096 & 0.6997 & \underline{0.9229} & 0.3401 & 0.6349 \\
Ours  & \underline{0.9054} & \underline{0.3152} & \textbf{0.9284}
        & \textbf{0.9238} & \underline{0.3485} & \textbf{0.8854} \\
\hline
\end{tabular}
}
\caption{Comparison of instruction-based editing methods on EmuEdit-bench and GEdit-bench with multiple metrics.}
\label{tab:edit_compare}
\end{table}

\begin{table}[ht]
\centering
\small
\setlength{\tabcolsep}{3pt} 
\renewcommand{\arraystretch}{1.1}
\resizebox{\columnwidth}{!}{
\begin{tabular}{l|ccccc|ccccc}
\hline
\multirow{2}{*}{\textbf{Method}} &
\multicolumn{5}{c|}{\textbf{DreamBench++}} &
\multicolumn{5}{c}{\textbf{OmniContext}} \\
\cline{2-11}
 & CLIP-cap & CLIP-ref & DINO-ref & Face-ref & Style-ref & CLIP-cap & CLIP-ref & DINO-ref & Face-ref & Style-ref \\
\hline
DreamO & 0.2899 & 0.7792 & 0.7518 & 0.335 & 0.5355 & 0.2986 & 0.7302 & 0.7075 & 0.4522 & - \\
Ominicontrol & 0.296 & 0.7642 & 0.6991 & 0.0579 & 0.3876 & 0.3067 & 0.7009 & 0.6126 & - & - \\
UNO & 0.2832 & 0.776 & 0.7429 & 0.2572 & 0.4328 & 0.2962 & 0.7106 & 0.6961 & 0.3665 & - \\
\hline
ACE++ & 0.2791 & 0.7759 & 0.732 & 0.1636 & 0.5306 & 0.2832 & 0.7183 & 0.6932 & 0.1789 & - \\
Kontext & 0.2829 & \textbf{0.819} & \underline{0.7919} & \underline{0.3429} & \underline{0.5655} & 0.2962 & \textbf{0.765} & 0.7494 & \underline{0.5596} & - \\
BAGEL & \textbf{0.3036} & 0.7338 & 0.6998 & 0.0487 & 0.5065 & 0.2914 & 0.7188 & 0.7094 & 0.1264 & - \\
OmniGen2 & 0.298 & 0.7712 & 0.752 & 0.1213 & 0.5167 & 0.3056 & \underline{0.7544} & 0.7289 & 0.3919 & - \\
Qwen-Edit & 0.3009 & 0.7595 & 0.7187 & 0.2188 & 0.5095 & \textbf{0.3152} & 0.7115 & 0.6797 & 0.3019 & - \\
DreamOmni2 & 0.2731 & \underline{0.8062} & \textbf{0.8008} & 0.2344 & 0.5364 & 0.2848 & 0.7733 & \underline{0.7611} & 0.5111 & - \\
Ours & \underline{0.3011} & 0.7906 & 0.7613 & \textbf{0.3678} & \textbf{0.5679} & \underline{0.3096} & 0.7297 & \textbf{0.7628} & \textbf{0.5607} & - \\
\hline
\end{tabular}
}
\caption{Comparison of subject-driven generation methods on DreamBench++ and OmniContext with multiple metrics.}
\label{tab:subject_compare}
\end{table}

\paragraph{Subject driven evaluation.}
We evaluate our model’s fine-grained preservation ability against specialized subject-driven generation methods on DreamBench++ and OmniContext, with results shown in Tab.~\ref{tab:subject_compare}. We focus on metrics that measure subject, identity, and style fidelity (noting that OmniContext does not include style-related tasks).
The results indicate strong preservation performance: our model achieves SOTA Face-ref scores on both benchmarks and the highest Style-ref score on DreamBench++. In addition, we obtain the top DINO-ref score on OmniContext and remain highly competitive on DreamBench++. These findings demonstrate that our unified model can match or surpass specialized models, effectively mitigating the typical tension between subject fidelity and generative diversity.

\begin{figure}[!h]
  \centering
  \includegraphics[width=0.95\linewidth]{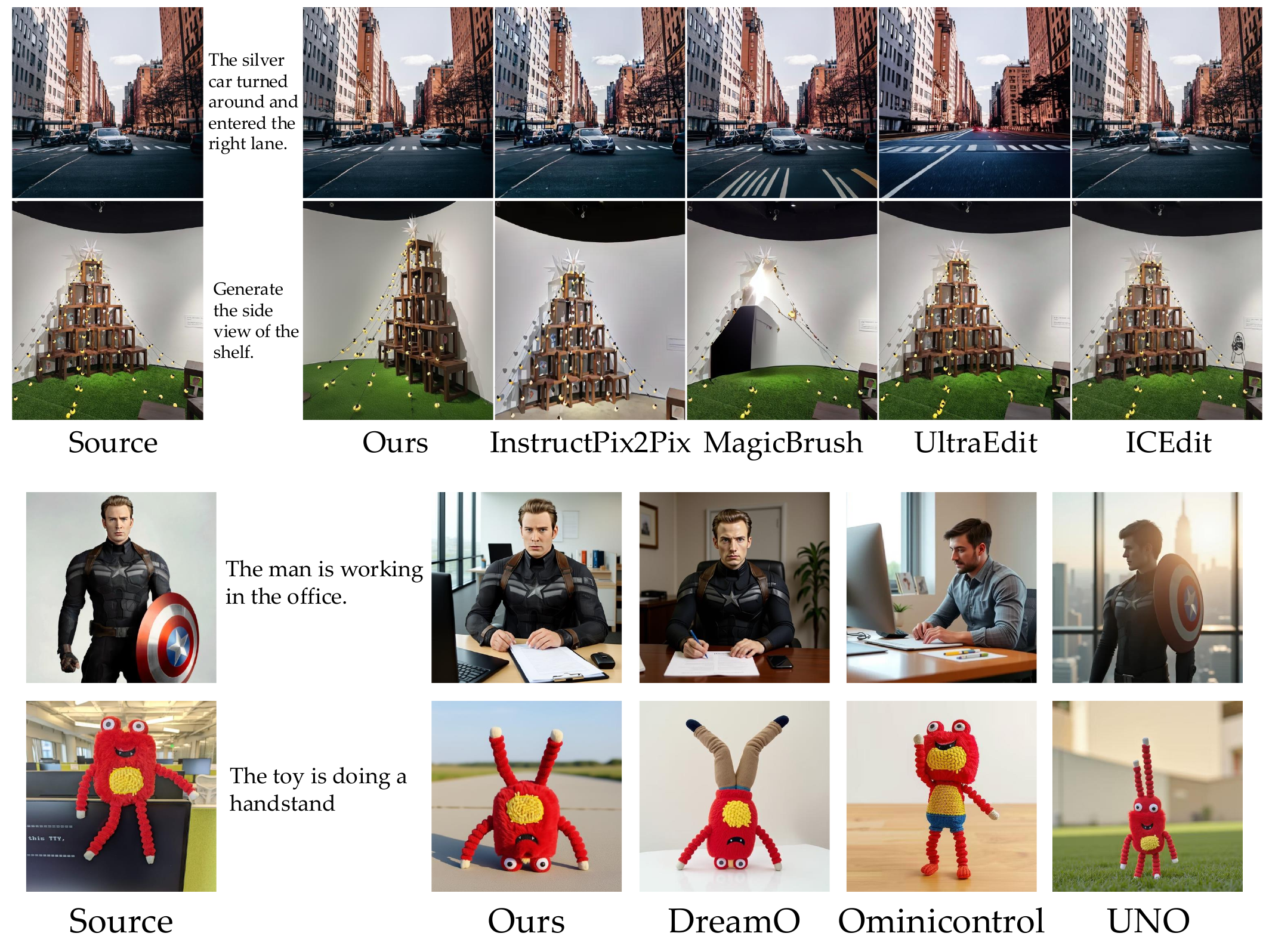}
  \caption{Compare with specialized image editing models and subject-driven generation models.}
  \vspace{-1em}
  \label{fig:compare_edit}
\end{figure}
\subsection{Qualitative Comparison}
\paragraph{Qualitative comparison with unified baselines.}
As demonstrated in the preceding qualitative comparison (Fig.~\ref{fig:main_compare}), our method consistently surpasses SOTA baselines in complex tasks characterized by interfering intents. 
These unified models typically fail to resolve inherent task conflicts, resulting in critical failures such as ``copy-paste'' artifacts in subject-driven generation, stylistic dissonance during inpainting, or incomplete execution in compositional editing. 
Our approach successfully navigates these challenges by utilizing the Predictive Alignment Regularization. This mechanism effectively decouples and routes conflicting sub-tasks (e.g. local semantic edits versus global style preservation) to specialized experts, thereby mitigating the core task interference that plagues unified models.

\paragraph{Qualitative comparison with specialized baselines.} We further present a comprehensive comparison against specialized image editing methods (InstructPix2Pix~\cite{brooks2023instructpix2pix}, MagicBrush~\cite{zhang2023magicbrush}, UltraEdit~\cite{zhao2024ultraedit}, ICEdit~\cite{zhang2025context}) and subject-driven models (DreamO~\cite{mou2025dreamo}, OmniControl~\cite{tan2025ominicontrol}, UNO~\cite{wu2025less}) in Fig.~\ref{fig:compare_edit}. For image editing, specialized baselines struggle with significant structural or geometric changes. As shown in the silver car case, they fail to execute the complex motion of turning around, resulting in minor texture changes; similarly, they fail to synthesize the side view of the complex shelf structure. In contrast, our method accurately handles these 3D-aware edits, benefiting from the structural diversity and geometric awareness implicitly learned from subject-driven data. Conversely, in subject-driven tasks, specialized models often compromise identity or instruction following. For the human subject, baselines either lose facial identity/clothing details (OmniControl) or fail to render the office context (UNO). For the toy subject requiring a handstand, baselines generate incorrect upright poses. Our method, however, maintains robust identity while adhering to complex motion instructions. This enhanced fidelity is attributed to the high consistency derived from editing alignment data during unified training. Overall, our model effectively handles both task types by leveraging semantic-aligned routing. This mechanism assigns conflicting objectives to specialized experts, enabling cross-task benefits: generative diversity from subject data improves editing geometry, while fidelity constraints from editing data enhance identity preservation in generation.

\subsection{Ablation Study}
\begin{figure*}[t]
  \centering
  \includegraphics[width=0.95\linewidth]{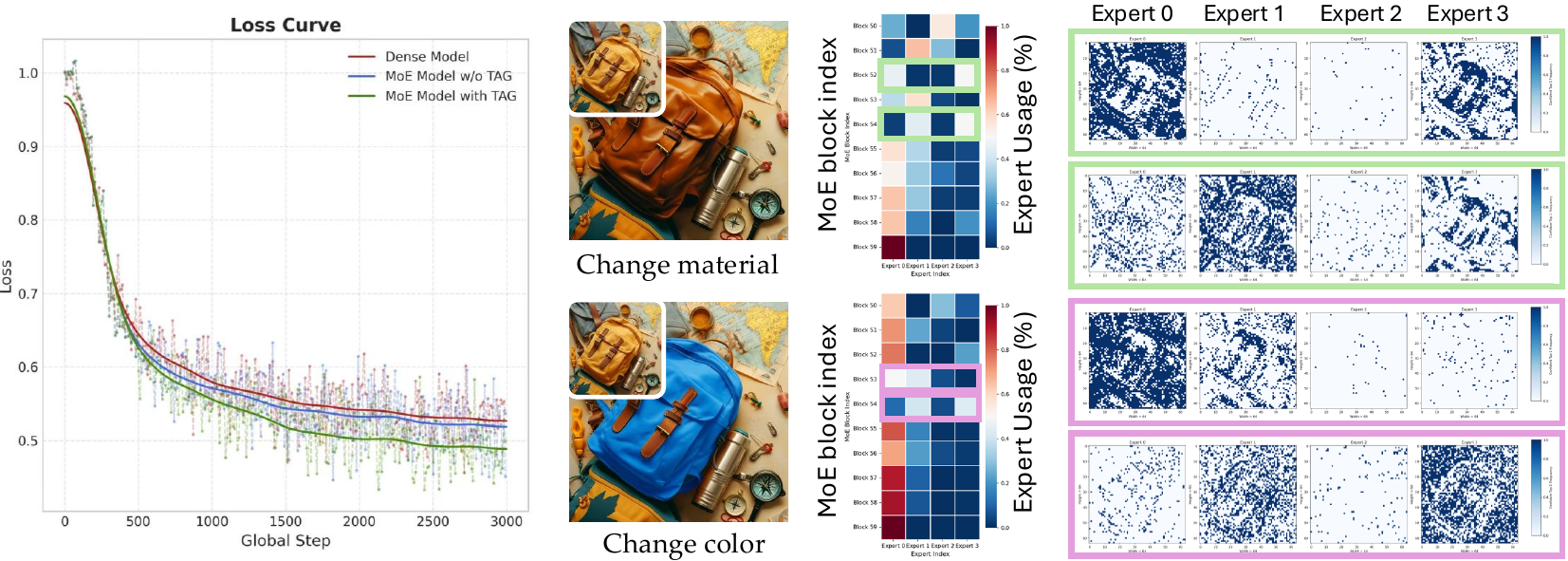}
  \caption{Left: Training loss curves of the dense and MoE architecture. Right: Token strategy in different generation tasks.}
  \label{fig:analysis_moe}
\end{figure*}

\paragraph{Effectiveness of the MoE architecture.} We compare our sparse MoE architecture to a dense baseline of an equivalent activated parameter count. This dense model shows a severe performance drop on ICE-Bench metrics (Tab.\ref{tab:ablation}) and slower convergence (Fig.~\ref{fig:analysis_moe} left). This validates that the sparse architecture is fundamentally more effective at mitigating the severe task interference inherent in the unified task space than a computationally-equivalent dense model.
\paragraph{Effect of predictive alignment regularization.}
We ablate the semantic-alignment loss by removing $\mathcal{L}_{align}$. Without this loss, the MoE gating network performs task-agnostic expert selection, receiving no semantic guidance from our hierarchical tags. As shown in Tab.~\ref{tab:ablation}, this variant exhibits substantial degradation across all major metrics.
This finding is key: a sparse MoE architecture alone is not sufficient. $\mathcal{L}_{align}$ is what enables semantically guided routing, which is essential for mitigating task interference.
Notably, the MoE w/o $\mathcal{L}_{align}$ variant still surpasses the dense baseline, benefiting from the larger effective capacity of the sparse MoE structure, which allows exploration of a richer solution space under the same computational budget.

\paragraph{Analysis of expert specialization.}
To provide direct evidence of our method's success, we visualize the inference-time routing decisions and analyze the internal expert activation patterns.
Our analysis is a two-step process. First, we compute an ``Expert Utilization Rate'' for each MoE layer (shown as the heatmap in the middle of Fig.~\ref{fig:analysis_moe}), which represents the percentage of total image tokens routed to each expert. A utilization of 0\% (blue) or 100\% (red) indicates no specialization. We focus our analysis on layers exhibiting differentiated routing, where utilization is mixed (near white), as this is where functional specialization occurs.
Second, for these active layers, we visualize the per-token routing scores for each expert, reshaping them to the image's spatial dimensions. In these token heatmaps, a high score (blue) indicates that the corresponding image tokens are strongly routed to that specific expert. The results reveal a clear, spatially-aware, and task-specific specialization. For Change Material and Change Color, the model activates distinct combinations of experts. Critically, the token heatmaps for these active experts show that computation is spatially concentrated on the backpack's pixels, precisely the region relevant to the edit. The non-relevant background tokens are correctly routed to other experts (or have near-zero activation for these experts).
This analysis provides strong evidence that our model has learned a sophisticated specialization that is both task-specific (using unique expert combinations for different tasks) and spatially-aware (experts learn to process semantically relevant image regions). This confirms our method successfully resolves task conflicts by dispatching them to distinct, specialized computational pathways.

\begin{table}[t]
\centering
\small
\setlength{\tabcolsep}{3pt}
\renewcommand{\arraystretch}{1.1}
\resizebox{\columnwidth}{!}{
\begin{tabular}{l|cccccccc}
\hline
\textcolor{white}{xxx}\textbf{Method}  &DINO-ref & Face-ref & Style-ref & CLIP-src  & CLIP-cap & vllmqa \\ \hline
Dense     &0.7196 &0.3544 &0.5177 &0.851   &0.263  &0.637  \\
MoE w/o $\mathcal{L}_{align}$  &0.7355 &0.3779 &0.5251  &0.863   &0.274  &0.677  \\
MoE w/ $\mathcal{L}_{align}$   &0.7620 &0.4642 &0.5679  &0.879   &0.281  &0.847 \\
\hline
\end{tabular}
}
\caption{Ablation study on dense model and predictive alignment regularization.}
\label{tab:ablation}
\end{table}

\subsection{User study}
\begin{figure}[!h]
  \centering
  \includegraphics[width=0.95\linewidth]{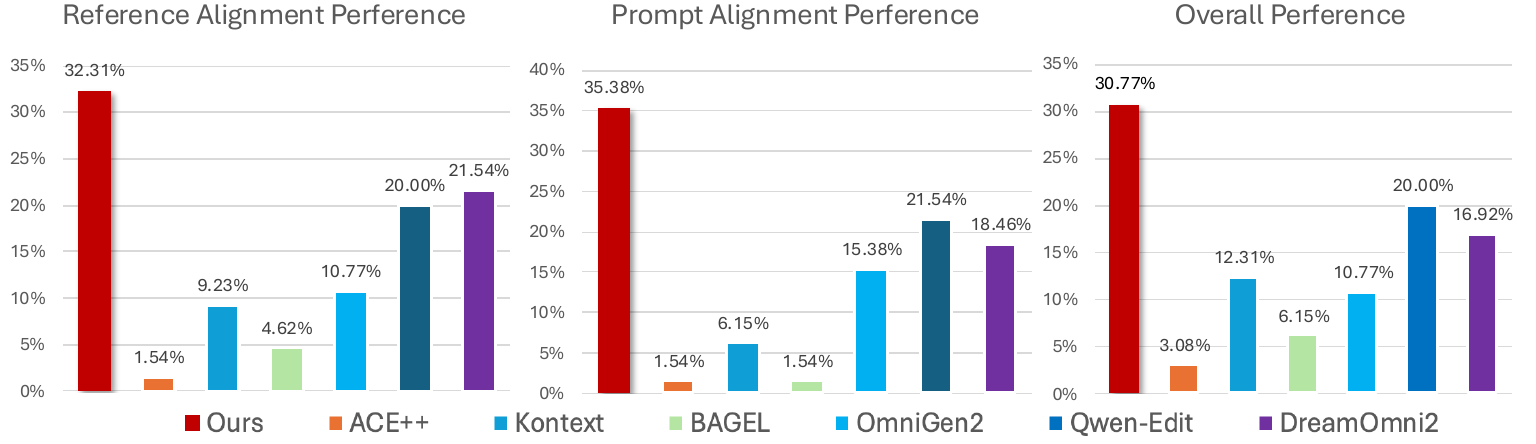}
  \caption{User study on reference alignment, prompt alignment and overall perference.}
  \label{fig:user_study}
\end{figure}

We conducted a user study with 65 participants on 50 cases from ICE-Bench~\cite{pan2025ice}. Participants were asked to select the single best result according to three criteria: (1) Reference Alignment (consistency with the source image), (2) Prompt Alignment (faithfulness to the textual instruction), and (3) Overall Preference (overall visual quality).
In total, 350 sets were evaluated, and the aggregated results are shown in Fig.~\ref{fig:user_study}. The results reveal a clear and consistent preference for our method, which achieved the highest selection rate across all three evaluation criteria.
\section{Limitations and Future Work}
A key limitation is our framework's lack of unified input understanding. Our model relies on pre-processed instructions (the intent) and cannot jointly reason over this intent and the visual content of the source image. This separation restricts tasks requiring integrated semantic and perceptual understanding.
For instance, our model fails at content-based reasoning (e.g., solving a math problem in an image) because it understands the editing intent (e.g., scope, type) but not the contextual information in the pixels themselves.
A promising future direction is an end-to-end system incorporating a multimodal reasoning engine to unify perceptual understanding (content), intent comprehension (command), and conceptual generation (reasoning).

\section{Conclusion}
\label{sec:conclusion}
In this paper, we propose TAG-MoE, a task-aware MoE framework for unified image generation and editing. We identify the task-agnostic routing as the core bottleneck for applying MoE to diverse, conflicting tasks. To address this, we introduce a Hierarchical Task Semantic Annotation scheme and Predictive Alignment regularization to effectively injects global task intent into the local routing decisions, forcing the model to develop meaningful expert specialization. Our experiments demonstrate that TAG-MoE significantly mitigates task interference, outperforming dense models and task-agnostic MoE baselines in both quantitative metrics and qualitative fidelity.

{
    \small
    \bibliographystyle{ieeenat_fullname}
    \bibliography{main}

@inproceedings{chen2025unireal,
  title={Unireal: Universal image generation and editing via learning real-world dynamics},
  author={Chen, Xi and Zhang, Zhifei and Zhang, He and Zhou, Yuqian and Kim, Soo Ye and Liu, Qing and Li, Yijun and Zhang, Jianming and Zhao, Nanxuan and Wang, Yilin and others},
  booktitle={Proceedings of the Computer Vision and Pattern Recognition Conference},
  pages={12501--12511},
  year={2025}
}

@inproceedings{xiao2025omnigen,
  title={Omnigen: Unified image generation},
  author={Xiao, Shitao and Wang, Yueze and Zhou, Junjie and Yuan, Huaying and Xing, Xingrun and Yan, Ruiran and Li, Chaofan and Wang, Shuting and Huang, Tiejun and Liu, Zheng},
  booktitle={Proceedings of the Computer Vision and Pattern Recognition Conference},
  pages={13294--13304},
  year={2025}
}

@article{wu2025omnigen2,
  title={OmniGen2: Exploration to Advanced Multimodal Generation},
  author={Wu, Chenyuan and Zheng, Pengfei and Yan, Ruiran and Xiao, Shitao and Luo, Xin and Wang, Yueze and Li, Wanli and Jiang, Xiyan and Liu, Yexin and Zhou, Junjie and others},
  journal={arXiv preprint arXiv:2506.18871},
  year={2025}
}

@article{deng2025emerging,
  title={Emerging properties in unified multimodal pretraining},
  author={Deng, Chaorui and Zhu, Deyao and Li, Kunchang and Gou, Chenhui and Li, Feng and Wang, Zeyu and Zhong, Shu and Yu, Weihao and Nie, Xiaonan and Song, Ziang and others},
  journal={arXiv preprint arXiv:2505.14683},
  year={2025}
}

@article{labs2025flux,
  title={FLUX. 1 Kontext: Flow Matching for In-Context Image Generation and Editing in Latent Space},
  author={Labs, Black Forest and Batifol, Stephen and Blattmann, Andreas and Boesel, Frederic and Consul, Saksham and Diagne, Cyril and Dockhorn, Tim and English, Jack and English, Zion and Esser, Patrick and others},
  journal={arXiv preprint arXiv:2506.15742},
  year={2025}
}

@inproceedings{han2025ace,
  title={ACE: All-round Creator and Editor Following Instructions via Diffusion Transformer},
  author={Han, Zhen and Jiang, Zeyinzi and Pan, Yulin and Zhang, Jingfeng and Mao, Chaojie and Xie, Chen-Wei and Liu, Yu and Zhou, Jingren},
  booktitle={The Thirteenth International Conference on Learning Representations},
  year={2025}
}

@article{fu2025univg,
  title={Univg: A generalist diffusion model for unified image generation and editing},
  author={Fu, Tsu-Jui and Qian, Yusu and Chen, Chen and Hu, Wenze and Gan, Zhe and Yang, Yinfei},
  journal={arXiv preprint arXiv:2503.12652},
  year={2025}
}

@article{lin2025realgeneral,
  title={Realgeneral: Unifying visual generation via temporal in-context learning with video models},
  author={Lin, Yijing and Huang, Mengqi and Zhuang, Shuhan and Mao, Zhendong},
  journal={arXiv preprint arXiv:2503.10406},
  year={2025}
}

@article{li2025blobctrl,
  title={Blobctrl: A unified and flexible framework for element-level image generation and editing},
  author={Li, Yaowei and Li, Lingen and Zhang, Zhaoyang and Li, Xiaoyu and Wang, Guangzhi and Li, Hongxiang and Cun, Xiaodong and Shan, Ying and Zou, Yuexian},
  journal={arXiv preprint arXiv:2503.13434},
  year={2025}
}

@article{fei2024scaling,
  title={Scaling diffusion transformers to 16 billion parameters},
  author={Fei, Zhengcong and Fan, Mingyuan and Yu, Changqian and Li, Debang and Huang, Junshi},
  journal={arXiv preprint arXiv:2407.11633},
  year={2024}
}

@article{cao2025hunyuanimage,
  title={HunyuanImage 3.0 Technical Report},
  author={Cao, Siyu and Chen, Hangting and Chen, Peng and Cheng, Yiji and Cui, Yutao and Deng, Xinchi and Dong, Ying and Gong, Kipper and Gu, Tianpeng and Gu, Xiusen and others},
  journal={arXiv preprint arXiv:2509.23951},
  year={2025}
}

@article{zhang2025context,
  title={In-context edit: Enabling instructional image editing with in-context generation in large scale diffusion transformer},
  author={Zhang, Zechuan and Xie, Ji and Lu, Yu and Yang, Zongxin and Yang, Yi},
  journal={arXiv preprint arXiv:2504.20690},
  year={2025}
}

@article{mao2025ace++,
  title={Ace++: Instruction-based image creation and editing via context-aware content filling},
  author={Mao, Chaojie and Zhang, Jingfeng and Pan, Yulin and Jiang, Zeyinzi and Han, Zhen and Liu, Yu and Zhou, Jingren},
  journal={arXiv preprint arXiv:2501.02487},
  year={2025}
}

@article{song2025query,
  title={Query-Kontext: An Unified Multimodal Model for Image Generation and Editing},
  author={Song, Yuxin and Dong, Wenkai and Wang, Shizun and Zhang, Qi and Xue, Song and Yuan, Tao and Yang, Hu and Feng, Haocheng and Zhou, Hang and Xiao, Xinyan and others},
  journal={arXiv preprint arXiv:2509.26641},
  year={2025}
}

@misc{deepseekai2024deepseekv3technicalreport,
      title={DeepSeek-V3 Technical Report}, 
      author={DeepSeek-AI},
      year={2024},
      eprint={2412.19437},
      archivePrefix={arXiv},
      primaryClass={cs.CL},
      url={https://arxiv.org/abs/2412.19437}, 
}

@inproceedings{rajbhandari2022deepspeed,
  title={Deepspeed-moe: Advancing mixture-of-experts inference and training to power next-generation ai scale},
  author={Rajbhandari, Samyam and Li, Conglong and Yao, Zhewei and Zhang, Minjia and Aminabadi, Reza Yazdani and Awan, Ammar Ahmad and Rasley, Jeff and He, Yuxiong},
  booktitle={International conference on machine learning},
  pages={18332--18346},
  year={2022},
  organization={PMLR}
}

@inproceedings{peebles2023scalable,
  title={Scalable diffusion models with transformers},
  author={Peebles, William and Xie, Saining},
  booktitle={Proceedings of the IEEE/CVF international conference on computer vision},
  pages={4195--4205},
  year={2023}
}

@article{wang2025seededit,
  title={SeedEdit 3.0: Fast and High-Quality Generative Image Editing},
  author={Wang, Peng and Shi, Yichun and Lian, Xiaochen and Zhai, Zhonghua and Xia, Xin and Xiao, Xuefeng and Huang, Weilin and Yang, Jianchao},
  journal={arXiv preprint arXiv:2506.05083},
  year={2025}
}

@article{bai2025qwen2,
  title={Qwen2. 5-vl technical report},
  author={Bai, Shuai and Chen, Keqin and Liu, Xuejing and Wang, Jialin and Ge, Wenbin and Song, Sibo and Dang, Kai and Wang, Peng and Wang, Shijie and Tang, Jun and others},
  journal={arXiv preprint arXiv:2502.13923},
  year={2025}
}

@article{zheng2025dense2moe,
  title={Dense2MoE: Restructuring Diffusion Transformer to MoE for Efficient Text-to-Image Generation},
  author={Zheng, Youwei and Ren, Yuxi and Xia, Xin and Xiao, Xuefeng and Xie, Xiaohua},
  journal={arXiv preprint arXiv:2510.09094},
  year={2025}
}

@article{wu2025qwen,
  title={Qwen-image technical report},
  author={Wu, Chenfei and Li, Jiahao and Zhou, Jingren and Lin, Junyang and Gao, Kaiyuan and Yan, Kun and Yin, Sheng-ming and Bai, Shuai and Xu, Xiao and Chen, Yilei and others},
  journal={arXiv preprint arXiv:2508.02324},
  year={2025}
}

@article{hurst2024gpt,
  title={Gpt-4o system card},
  author={Hurst, Aaron and Lerer, Adam and Goucher, Adam P and Perelman, Adam and Ramesh, Aditya and Clark, Aidan and Ostrow, AJ and Welihinda, Akila and Hayes, Alan and Radford, Alec and others},
  journal={arXiv preprint arXiv:2410.21276},
  year={2024}
}

@article{comanici2025gemini,
  title={Gemini 2.5: Pushing the frontier with advanced reasoning, multimodality, long context, and next generation agentic capabilities},
  author={Comanici, Gheorghe and Bieber, Eric and Schaekermann, Mike and Pasupat, Ice and Sachdeva, Noveen and Dhillon, Inderjit and Blistein, Marcel and Ram, Ori and Zhang, Dan and Rosen, Evan and others},
  journal={arXiv preprint arXiv:2507.06261},
  year={2025}
}

@inproceedings{caron2021emerging,
  title={Emerging properties in self-supervised vision transformers},
  author={Caron, Mathilde and Touvron, Hugo and Misra, Ishan and J{\'e}gou, Herv{\'e} and Mairal, Julien and Bojanowski, Piotr and Joulin, Armand},
  booktitle={Proceedings of the IEEE/CVF international conference on computer vision},
  pages={9650--9660},
  year={2021}
}

@misc{deepinsight2021insightface,
  author       = {{deepinsight}},
  title        = {InsightFace},
  howpublished = {\url{https://github.com/deepinsight/insightface}},
  year         = {2021},
  note         = {Accessed: 2025-11-04}
}

@article{pan2025ice,
  title={Ice-bench: A unified and comprehensive benchmark for image creating and editing},
  author={Pan, Yulin and He, Xiangteng and Mao, Chaojie and Han, Zhen and Jiang, Zeyinzi and Zhang, Jingfeng and Liu, Yu},
  journal={arXiv preprint arXiv:2503.14482},
  year={2025}
}

@inproceedings{sheynin2024emu,
  title={Emu edit: Precise image editing via recognition and generation tasks},
  author={Sheynin, Shelly and Polyak, Adam and Singer, Uriel and Kirstain, Yuval and Zohar, Amit and Ashual, Oron and Parikh, Devi and Taigman, Yaniv},
  booktitle={Proceedings of the IEEE/CVF Conference on Computer Vision and Pattern Recognition},
  pages={8871--8879},
  year={2024}
}

@article{liu2025step1x,
  title={Step1x-edit: A practical framework for general image editing},
  author={Liu, Shiyu and Han, Yucheng and Xing, Peng and Yin, Fukun and Wang, Rui and Cheng, Wei and Liao, Jiaqi and Wang, Yingming and Fu, Honghao and Han, Chunrui and others},
  journal={arXiv preprint arXiv:2504.17761},
  year={2025}
}

@inproceedings{peng2024dreambench,
  author={Yuang Peng and Yuxin Cui and Haomiao Tang and Zekun Qi and Runpei Dong and Jing Bai and Chunrui Han and Zheng Ge and Xiangyu Zhang and Shu-Tao Xia},
  title={DreamBench++: A Human-Aligned Benchmark for Personalized Image Generation},
  booktitle={The Thirteenth International Conference on Learning Representations},
  url={https://openreview.net/forum?id=4GSOESJrk6},
  year={2025},
}

@article{wang2024qwen2,
  title={Qwen2-vl: Enhancing vision-language model's perception of the world at any resolution},
  author={Wang, Peng and Bai, Shuai and Tan, Sinan and Wang, Shijie and Fan, Zhihao and Bai, Jinze and Chen, Keqin and Liu, Xuejing and Wang, Jialin and Ge, Wenbin and others},
  journal={arXiv preprint arXiv:2409.12191},
  year={2024}
}

@online{openai2025gpt4o,
  author  = {OpenAI},
  title   = {GPT-4o},
  year    = {2025},
  url     = {https://openai.com/index/introducing-4o-image-generation},
  urldate = {2025-02-05}
}

@article{somepalli2024measuring,
  title={Measuring style similarity in diffusion models},
  author={Somepalli, Gowthami and Gupta, Anubhav and Gupta, Kamal and Palta, Shramay and Goldblum, Micah and Geiping, Jonas and Shrivastava, Abhinav and Goldstein, Tom},
  journal={arXiv preprint arXiv:2404.01292},
  year={2024}
}

@techreport{google2025nanobanana,
  author      = {Google},
  title       = {Nano Banana},
  institution = {Google},
  year        = {2025},
  type        = {Technical Report}
}

@inproceedings{brooks2023instructpix2pix,
  title={Instructpix2pix: Learning to follow image editing instructions},
  author={Brooks, Tim and Holynski, Aleksander and Efros, Alexei A},
  booktitle={Proceedings of the IEEE/CVF conference on computer vision and pattern recognition},
  pages={18392--18402},
  year={2023}
}

@article{zhao2024ultraedit,
  title={Ultraedit: Instruction-based fine-grained image editing at scale},
  author={Zhao, Haozhe and Ma, Xiaojian Shawn and Chen, Liang and Si, Shuzheng and Wu, Rujie and An, Kaikai and Yu, Peiyu and Zhang, Minjia and Li, Qing and Chang, Baobao},
  journal={Advances in Neural Information Processing Systems},
  volume={37},
  pages={3058--3093},
  year={2024}
}

@inproceedings{choi2021viton,
  title={Viton-hd: High-resolution virtual try-on via misalignment-aware normalization},
  author={Choi, Seunghwan and Park, Sunghyun and Lee, Minsoo and Choo, Jaegul},
  booktitle={Proceedings of the IEEE/CVF conference on computer vision and pattern recognition},
  pages={14131--14140},
  year={2021}
}

@inproceedings{tan2025ominicontrol,
  title={Ominicontrol: Minimal and universal control for diffusion transformer},
  author={Tan, Zhenxiong and Liu, Songhua and Yang, Xingyi and Xue, Qiaochu and Wang, Xinchao},
  booktitle={Proceedings of the IEEE/CVF International Conference on Computer Vision},
  pages={14940--14950},
  year={2025}
}

@inproceedings{wei2024omniedit,
  title={Omniedit: Building image editing generalist models through specialist supervision},
  author={Wei, Cong and Xiong, Zheyang and Ren, Weiming and Du, Xeron and Zhang, Ge and Chen, Wenhu},
  booktitle={The Thirteenth International Conference on Learning Representations},
  year={2024}
}

@inproceedings{zhang2023adding,
  title={Adding conditional control to text-to-image diffusion models},
  author={Zhang, Lvmin and Rao, Anyi and Agrawala, Maneesh},
  booktitle={Proceedings of the IEEE/CVF international conference on computer vision},
  pages={3836--3847},
  year={2023}
}

@article{xia2025dreamomni2,
  title={DreamOmni2: Multimodal Instruction-based Editing and Generation},
  author={Xia, Bin and Peng, Bohao and Zhang, Yuechen and Huang, Junjia and Liu, Jiyang and Li, Jingyao and Tan, Haoru and Wu, Sitong and Wang, Chengyao and Wang, Yitong and others},
  journal={arXiv preprint arXiv:2510.06679},
  year={2025}
}

@article{zhang2023magicbrush,
  title={Magicbrush: A manually annotated dataset for instruction-guided image editing},
  author={Zhang, Kai and Mo, Lingbo and Chen, Wenhu and Sun, Huan and Su, Yu},
  journal={Advances in Neural Information Processing Systems},
  volume={36},
  pages={31428--31449},
  year={2023}
}

@article{mou2025dreamo,
  title={Dreamo: A unified framework for image customization},
  author={Mou, Chong and Wu, Yanze and Wu, Wenxu and Guo, Zinan and Zhang, Pengze and Cheng, Yufeng and Luo, Yiming and Ding, Fei and Zhang, Shiwen and Li, Xinghui and others},
  journal={arXiv preprint arXiv:2504.16915},
  year={2025}
}

@article{a,
  title={Less-to-more generalization: Unlocking more controllability by in-context generation},
  author={Wu, Shaojin and Huang, Mengqi and Wu, Wenxu and Cheng, Yufeng and Ding, Fei and He, Qian},
  journal={arXiv preprint arXiv:2504.02160},
  year={2025}
}

@article{xu2024headrouter,
  title={Headrouter: A training-free image editing framework for mm-dits by adaptively routing attention heads},
  author={Xu, Yu and Tang, Fan and Cao, Juan and Zhang, Yuxin and Kong, Xiaoyu and Li, Jintao and Deussen, Oliver and Lee, Tong-Yee},
  journal={arXiv preprint arXiv:2411.15034},
  year={2024}
}

@article{xu2025b4m,
  title={B4M: Breaking Low-Rank Adapter for Making Content-Style Customization},
  author={Xu, Yu and Tang, Fan and Cao, Juan and Zhang, Yuxin and Deussen, Oliver and Dong, Weiming and Li, Jintao and Lee, Tong-Yee},
  journal={ACM Transactions on Graphics},
  volume={44},
  number={2},
  pages={1--17},
  year={2025},
  publisher={ACM New York, NY}
}

@inproceedings{xu2025context,
  title={In-Context Brush: Zero-shot Customized Subject Insertion with Context-Aware Latent Space Manipulation},
  author={Xu, Yu and Tang, Fan and Wu, You and Gao, Lin and Deussen, Oliver and Yan, Hongbin and Li, Jintao and Cao, Juan and Lee, Tong-Yee},
  booktitle={Proceedings of the SIGGRAPH Asia 2025 Conference Papers},
  pages={1--12},
  year={2025}
}

@article{wu2025less,
  title={Less-to-more generalization: Unlocking more controllability by in-context generation},
  author={Wu, Shaojin and Huang, Mengqi and Wu, Wenxu and Cheng, Yufeng and Ding, Fei and He, Qian},
  journal={arXiv preprint arXiv:2504.02160},
  year={2025}
}

@inproceedings{wang2025ladb,
  title={LADB: Latent Aligned Diffusion Bridges for Semi-Supervised Domain Translation},
  author={Wang, Xuqin and Wu, Tao and Zhang, Yanfeng and Liu, Lu and Wang, Dong and Sun, Mingwei and Wang, Yongliang and Zeller, Niclas and Cremers, Daniel},
  booktitle={DAGM German Conference on Pattern Recognition},
  pages={221--236},
  year={2025},
  organization={Springer}
}

@article{wang2026geodesicnvs,
  title={GeodesicNVS: Probability Density Geodesic Flow Matching for Novel View Synthesis},
  author={Wang, Xuqin and Wu, Tao and Zhang, Yanfeng and Liu, Lu and Sun, Mingwei and Wang, Yongliang and Zeller, Niclas and Cremers, Daniel},
  journal={arXiv preprint arXiv:2603.01010},
  year={2026}
}

@inproceedings{zhu2025kv,
  title={Kv-edit: Training-free image editing for precise background preservation},
  author={Zhu, Tianrui and Zhang, Shiyi and Shao, Jiawei and Tang, Yansong},
  booktitle={Proceedings of the IEEE/CVF International Conference on Computer Vision},
  pages={16607--16617},
  year={2025}
}

@inproceedings{he2024customize,
  title={Customize your nerf: Adaptive source driven 3d scene editing via local-global iterative training},
  author={He, Runze and Huang, Shaofei and Nie, Xuecheng and Hui, Tianrui and Liu, Luoqi and Dai, Jiao and Han, Jizhong and Li, Guanbin and Liu, Si},
  booktitle={Proceedings of the IEEE/CVF conference on computer vision and pattern recognition},
  pages={6966--6975},
  year={2024}
}

@inproceedings{ruiz2023dreambooth,
  title={Dreambooth: Fine tuning text-to-image diffusion models for subject-driven generation},
  author={Ruiz, Nataniel and Li, Yuanzhen and Jampani, Varun and Pritch, Yael and Rubinstein, Michael and Aberman, Kfir},
  booktitle={Proceedings of the IEEE/CVF conference on computer vision and pattern recognition},
  pages={22500--22510},
  year={2023}
}

@article{xu2025attribute,
  title={Attribute guided adversarial editing for face privacy protection},
  author={Xu, Yu and Wang, Ziang and Tang, Fan and Cao, Juan and Li, Xirong and Li, Jintao},
  journal={Visual Informatics},
  pages={100267},
  year={2025},
  publisher={Elsevier}
}

@article{xu2026beyond,
  title={Beyond Pixels: Visual Metaphor Transfer via Schema-Driven Agentic Reasoning},
  author={Xu, Yu and Zhang, Yuxin and Cao, Juan and Gao, Lin and Wang, Chunyu and Deussen, Oliver and Lee, Tong-Yee and Tang, Fan},
  journal={arXiv preprint arXiv:2602.01335},
  year={2026}
}

@article{yin2026generative,
  title={Generative Visual Chain-of-Thought for Image Editing},
  author={Yin, Zijin and Hang, Tiankai and Cheng, Yiji and Zhang, Shiyi and He, Runze and Xu, Yu and Wang, Chunyu and Li, Bing and Chang, Zheng and Liang, Kongming and others},
  journal={arXiv preprint arXiv:2603.01893},
  year={2026}
}

@article{he2026re,
  title={Re-Align: Structured Reasoning-guided Alignment for In-Context Image Generation and Editing},
  author={He, Runze and Cheng, Yiji and Hang, Tiankai and Li, Zhimin and Xu, Yu and Yin, Zijin and Zhang, Shiyi and Dai, Wenxun and Du, Penghui and Ma, Ao and others},
  journal={arXiv preprint arXiv:2601.05124},
  year={2026}
}

@inproceedings{yang2025magic,
  title={Magic-vqa: Multimodal and grounded inference with commonsense knowledge for visual question answering},
  author={Yang, Shuo and Han, Caren and Luo, Siwen and Hovy, Eduard},
  booktitle={Findings of the Association for Computational Linguistics: ACL 2025},
  pages={16967--16986},
  year={2025}
}

@inproceedings{dai2025latent,
  title={Latent Swap Joint Diffusion for 2D Long-Form Latent Generation},
  author={Dai, Yusheng and Wang, Chenxi and Li, Chang and Wang, Chen and Li, Kewei and Du, Jun and Sun, Lei and Gao, Jianqing and Wang, Ruoyu and Ma, Jiefeng},
  booktitle={Proceedings of the IEEE/CVF International Conference on Computer Vision},
  pages={11006--11015},
  year={2025}
}

@article{dai2026omni2sound,
  title={Omni2Sound: Towards Unified Video-Text-to-Audio Generation},
  author={Dai, Yusheng and Chen, Zehua and Jiang, Yuxuan and Gao, Baolong and Ke, Qiuhong and Zhu, Jun and Cai, Jianfei},
  journal={arXiv preprint arXiv:2601.02731},
  year={2026}
}
}


\end{document}